\documentclass[conference]{IEEEtran}
\IEEEoverridecommandlockouts
\usepackage{cite}
\usepackage{amsmath,amssymb,amsfonts}
\usepackage{algorithmic}
\usepackage{graphicx}
\usepackage{textcomp}
\usepackage{xcolor}
\def\BibTeX{{\rm B\kern-.05em{\sc i\kern-.025em b}\kern-.08em
    T\kern-.1667em\lower.7ex\hbox{E}\kern-.125emX}}
    
\usepackage{multirow}
\usepackage[
singlelinecheck=false % <-- important
]{caption}
\usepackage[justification=centering]{caption}

\begin{document}

\title{Natural language processing for clusterization of genes according to their functions\\

\thanks{This work was supported by Priority-2030 program, grant \#075-15-2021-1331.}
}

\author{\IEEEauthorblockN{Vladislav Dordiuk$^{1, 2}$}
\IEEEauthorblockA{\textit{$^{1}$Institute of Immunology and Physiology} \\
\textit{$^{2}$Ural Federal University}\\
Ekaterinburg, Russia \\
vladislav0860@gmail.com}
\and
\IEEEauthorblockN{Ekaterina Demicheva$^{1, 2}$}
\IEEEauthorblockA{\textit{$^{1}$Institute of Immunology and Physiology} \\
\textit{$^{2}$Ural Federal University}\\
Ekaterinburg, Russia \\
0000-0003-0744-5494}
\and
\IEEEauthorblockN{Fernando Polanco Espino$^{2}$}
\IEEEauthorblockA{\textit{$^{2}$Ural Federal University} \\
%\textit{Ural Federal University}\\
Ekaterinburg, Russia \\
fpolancoespino@gmail.com}

\and
%\linebreakand % <------------- \and with a line-break
\IEEEauthorblockN{Konstantin Ushenin$^{1, 2}$}
\IEEEauthorblockA{\textit{$^{1}$Institute of Immunology and Physiology} \\
\textit{$^{2}$Ural Federal University}\\
Ekaterinburg, Russia \\
0000-0003-0575-3506}
}

\maketitle

\begin{abstract}
There are hundreds of methods for analysis of data obtained in mRNA-sequencing. The most of them are focused on small number of genes. In this study, we propose an approach that reduces the analysis of several thousand genes to analysis of several clusters. The list of genes is enriched with information from open databases. Then, the descriptions are encoded as vectors using the pretrained language model (BERT) and some text processing approaches. The encoded gene function pass through the dimensionality reduction and clusterization. Aiming to find the most efficient pipeline, 180 cases of pipeline with different methods in the major pipeline steps were analyzed. The performance was evaluated with clusterization indexes and expert review of the results.
\end{abstract}

\begin{IEEEkeywords}
natural language processing, BERT, semantic analysis, differential gene expression analysis, gene ontology, gene expression, clusterization
\end{IEEEkeywords}

\section{Introduction}
mRNA-seq and differential gene expression analysis are powerful tools for the analysis of complex biological systems. They reveal complex intracellular processes and help to detect targets for new drugs. Differential gene expression analysis is also used for study of orphan and rare genetic diseases. During the last two decades, bioinformatics has adapted or invented many methods and software tools for the analysis, such as principal component analysis, agglomerative clusterization, Venn diagram \cite{venn}. The development of open access databases led to the invention of methods that utilize open information for the analysis, such as signaling pathways analysis \cite{pathways, pathways2}, protein-protein interaction, and others\cite{ppi}. Usually such methods use graphs for data representation and solve some optimization problems. 

The major shortcoming of mentioned approaches is focus on hundreds of genes. The bigger set of genes requires an extraordinary amount of work for the analysis of biological processes. The recent advances in natural language processing opened new opportunities for analysis of differential gene expression data. This work is aimed to create an automated method for differential gene expression analysis that uses pretrained language models pretrained. The work present the main pipeline for processing and estimates 180 modification of this pipeline. Estimation was performed with quantitative analysis of some clusterization metrics, and with an expert review. 

\section{Methods}

The proposed method was applied to big gene set with 12927 gene names. This list of names was obtained using mRNA differential gene expression analysis of an open dataset (GSE40561; \cite{GEOdataset}), that includes samples of healthy volunteers and patients with three autoinflammatory diseases: RBCK1 (HOIL1), NEMO, and MYD88 deficiencies. mRNA were extracted from whole blood and fibroblasts with Illumina HumanHT-12 V3.0 expression beadchip platform. The bioinformatic analysis includes several standard steps \cite{Stafford2007}: normalization of gene expression matrix, an empirical Bayes moderated t-test and subsequent volcano plot analysis.

In the first step of the proposed method, the list of genes was enriched with gene annotations. Then information about gene functions were processed with three approaches. The first approach utilizes data from a manually curated database and encodes information as a binary vector. The second approach uses regular expressions for processing of gene descriptions and, also, encodes information as a binary vector. The third approach uses linguistic models and transforms gene description as a vector in the semantic vector space.

A manually curated functional annotations were obtained from the Gene Ontology database. This approach provided 14701 GO-terms relating to one of the three categories: cellular component, molecular function, or biological process. For the following processing, these GO-term were encoded as binary vectors for each gene. True values show presence of GO-term in the gene description, and The false values shows the absence of GO-term in the gene descriptions. Thus, the vector consists of 14701 component.

A text description of the gene functions was obtained from UniProt database. Biological pathways and intracellular systems are mentioned in the text descriptions as acronyms. For instance, there are MYD88, NFKB1, GTPases, etc. All acronyms were mined from the text with a regular expression ('[a-z0-9]{0,3}[A-Z]{2,3}[a-z0-9]{0,3}'). Such approach has provided 5289 acronyms that were also encoded as binary vector.

Text descriptions of the genes also were transformed to vectors in the semantic space using a linguistic model. Transformation was performed with the BERT model that is a deep neural network with a bidirectional transformer architecture. We used six BERT model modifications that were pretrained on large text corpuses \cite{bert}. There are base BERT\cite{bert},  RoBERTa \cite{roberta},  Clinical BERT \cite{clinicalBERT}, BioBERT \cite{bioBERT}, BlueBERT pretrained on PubMed abstracts \cite{blueBERT}, and BlueBERT pretrained on PubMed abstracts and MIMIC-III \cite{blueBERT}.

% картинка с пайплайном
\begin{figure*}[ht]
\includegraphics[width=0.9\textwidth]{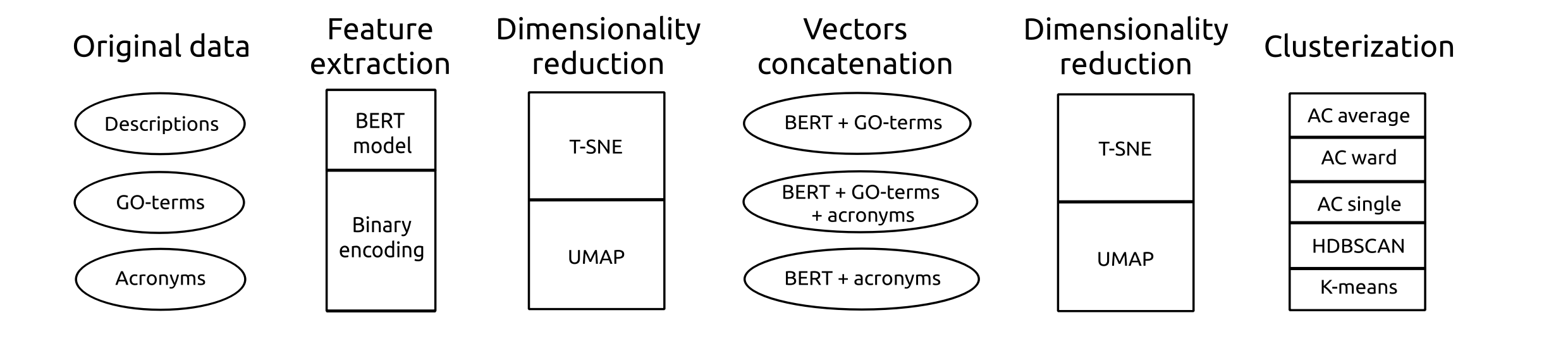}
\centering
\caption{Stages of the pipeline for semantic analysis of gene sets. The dimensionality reduction method in the second block is the same as in the first one.}
\label{pipeFig}
\end{figure*}

Each gene was encoded in three vectors: semantic of the text descriptions (3072 parameters), biological function from manually curated database (14701 parameters), and biological pathways mentioned in the text description (5289 parameters). Aiming to find the better processing pipeline, this study tried various combination of the vectors and methods of their analysis. The general design of study is shown in Fig. \ref{pipeFig}. The pipeline consists of the preprocessing, a dimensionality reduction, vector concatenation, one more dimensionality reduction, and clusterization.

In the first step, dimensionality of the vectors was reduced down to 50 components with one of non-linear methods of dimensionality reduction: t-SNE or UMAP\cite{umap}. Then, vectors were concatenated together in several combinations. There are three variants of the vector concatenation: the text descriptions and biological functions, the text descriptions and pathways, the text descriptions with biological functions and pathways. Each combination of vectors was passed through one more dimensionality reduction down to two components with the same method that was used in the previous time. Low dimensional representation of genes were clustered using k-means, HDBSCAN or the agglomerative clustering. The last one were applied with several approaches to the cluster linkage: the single-linkage, average-linkage or Ward's method \cite{clustering}. 

In total, several options for each stage of the pipeline provide experimental design with 180 computational experiments. That includes one of six language model, one of two dimensionality reduction methods, one of three combination of vectors,  and one of five clustering methods  (180 = 6 * 2 * 3 * 5). For qualitative interpretation of the results, we employ the Silhouette score \cite{sil_score} that assesses the quality of clusterization. It ranges from -1 to 1, where a high positive value means better clustering. After choosing the best three variants of pipeline according to the highest Silhouette scores. An expert manually changed hyperparameters of methods to reach the most clear and useful clusterization of the gene set, and conducted manual analysis of clusters to choose the pipeline that performs the best. The chosen variant was further tested on datasets for each disease individually. 

% таблица с результатами
\begingroup
\setlength{\tabcolsep}{10pt} % Default value: 6pt
\renewcommand{\arraystretch}{1.1} % Default value: 1
\begin{table*}[h]
\centering
\scriptsize
\begin{tabular}{cccccccc}
\multicolumn{1}{l}{}                                                                      & \multicolumn{1}{l}{}                                                                          & \multicolumn{2}{c}{Acronyms}                                                                                                                                      & \multicolumn{2}{c}{GO-terms}                                                                                                                                      & \multicolumn{2}{c}{GO-terms with acronyms}                                                                                                                        \\ \hline
Names                                                                                     & Clustering method                                                                             & T-SNE                                                                           & UMAP                                                                            & T-SNE                                                                           & UMAP                                                                            & T-SNE                                                                           & UMAP                                                                            \\ \hline
BaseBERT                                                                                  & \begin{tabular}[c]{@{}c@{}}AC single \\ AC ward\\ AC average\\ HDBSCAN\\ K-means\end{tabular} & \begin{tabular}[c]{@{}c@{}}-0.441\\ 0.338\\ 0.324\\ -0.315\\ 0.372\end{tabular} & \begin{tabular}[c]{@{}c@{}}-0.214\\ 0.502\\ 0.475\\ -0.059\\ 0.511\end{tabular} & \begin{tabular}[c]{@{}c@{}}-0.570\\ 0.325\\ 0.325\\ -0.446\\ 0.361\end{tabular} & \begin{tabular}[c]{@{}c@{}}-0.149\\ 0.458\\ 0.468\\ -0.139\\ 0.477\end{tabular} & \begin{tabular}[c]{@{}c@{}}-0.575\\ 0.302\\ 0.299\\ -0.607\\ 0.354\end{tabular} & \begin{tabular}[c]{@{}c@{}}-0.147\\ 0.500\\ 0.491\\ -0.178\\ 0.509\end{tabular} \\ \hline
RoBERTa                                                                                   & \begin{tabular}[c]{@{}c@{}}AC single \\ AC ward\\ AC average\\ HDBSCAN\\ K-means\end{tabular} & \begin{tabular}[c]{@{}c@{}}-0.565\\ 0.327\\ 0.329\\ -0.319\\ 0.366\end{tabular} & \begin{tabular}[c]{@{}c@{}}-0.041\\ 0.481\\ 0.479\\ -0.028\\ 0.495\end{tabular} & \begin{tabular}[c]{@{}c@{}}-0.527\\ 0.320\\ 0.332\\ -0.412\\ \textbf{0.367}\end{tabular} & \begin{tabular}[c]{@{}c@{}}0.190\\ 0.481\\ 0.483\\ 0.124\\ 0.491\end{tabular}   & \begin{tabular}[c]{@{}c@{}}-0.593\\ 0.292\\ 0.287\\ -0.666\\ 0.354\end{tabular} & \begin{tabular}[c]{@{}c@{}}-0.166\\ 0.505\\ 0.438\\ -0.149\\ 0.503\end{tabular} \\ \hline
BioBERT                                                                                   & \begin{tabular}[c]{@{}c@{}}AC single \\ AC ward\\ AC average\\ HDBSCAN\\ K-means\end{tabular} & \begin{tabular}[c]{@{}c@{}}-0.344\\ 0.329\\ 0.337\\ -0.348\\ 0.372\end{tabular} & \begin{tabular}[c]{@{}c@{}}-0.076\\ 0.513\\ 0.515\\ 0.109\\ 0.514\end{tabular}  & \begin{tabular}[c]{@{}c@{}}-0.614\\ 0.322\\ 0.320\\ -0.497\\ 0.361\end{tabular} & \begin{tabular}[c]{@{}c@{}}-0.048\\ 0.475\\ 0.454\\ -0.024\\ 0.494\end{tabular} & \begin{tabular}[c]{@{}c@{}}-0.632\\ 0.295\\ 0.294\\ -0.604\\ 0.355\end{tabular} & \begin{tabular}[c]{@{}c@{}}-0.226\\ 0.491\\ 0.472\\ -0.245\\ 0.477\end{tabular} \\ \hline
ClinicBERT                                                                                & \begin{tabular}[c]{@{}c@{}}AC single \\ AC ward\\ AC average\\ HDBSCAN\\ K-means\end{tabular} & \begin{tabular}[c]{@{}c@{}}-0.534\\ 0.326\\ 0.334\\ -0.435\\ 0.366\end{tabular} & \begin{tabular}[c]{@{}c@{}}0.057\\ 0.519\\ 0.505\\ -0.059\\ 0.523\end{tabular}  & \begin{tabular}[c]{@{}c@{}}-0.622\\ 0.317\\ 0.323\\ -0.480\\ 0.366\end{tabular} & \begin{tabular}[c]{@{}c@{}}-0.241\\ 0.481\\ 0.481\\ -0.001\\ 0.491\end{tabular} & \begin{tabular}[c]{@{}c@{}}-0.582\\ 0.303\\ 0.288\\ -0.586\\ 0.352\end{tabular} & \begin{tabular}[c]{@{}c@{}}-0.136\\ 0.509\\ 0.503\\ -0.079\\ 0.509\end{tabular} \\ \hline
\begin{tabular}[c]{@{}c@{}}BlueBERT \\ pretrained on PubMed\end{tabular}                  & \begin{tabular}[c]{@{}c@{}}AC single \\ AC ward\\ AC average\\ HDBSCAN\\ K-means\end{tabular} & \begin{tabular}[c]{@{}c@{}}-0.517\\ 0.325\\ 0.319\\ -0.364\\ 0.368\end{tabular} & \begin{tabular}[c]{@{}c@{}}-0.159\\ 0.520\\ 0.503\\ -0.092\\ 0.519\end{tabular} & \begin{tabular}[c]{@{}c@{}}-0.549\\ 0.320\\ 0.324\\ -0.459\\ 0.364\end{tabular} & \begin{tabular}[c]{@{}c@{}}-0.126\\ 0.480\\ 0.452\\ -0.029\\ \textbf{0.500}\end{tabular} & \begin{tabular}[c]{@{}c@{}}-0.635\\ 0.314\\ 0.311\\ -0.611\\ \textbf{0.362}\end{tabular} & \begin{tabular}[c]{@{}c@{}}-0.147\\ 0.515\\ 0.481\\ -0.025\\ \textbf{0.516}\end{tabular} \\ \hline
\begin{tabular}[c]{@{}c@{}}BlueBERT \\ pretrained on \\ PubMed and MIMIC-III\end{tabular} & \begin{tabular}[c]{@{}c@{}}AC single \\ AC ward\\ AC average\\ HDBSCAN\\ K-means\end{tabular} & \begin{tabular}[c]{@{}c@{}}-0.575\\ 0.342\\ 0.340\\ -0.404\\ \textbf{0.377}\end{tabular} & \begin{tabular}[c]{@{}c@{}}-0.149\\ 0.536\\ 0.521\\ -0.255\\ \textbf{0.546}\end{tabular} & \begin{tabular}[c]{@{}c@{}}-0.539\\ 0.323\\ 0.316\\ -0.526\\ 0.364\end{tabular} & \begin{tabular}[c]{@{}c@{}}-0.101\\ 0.461\\ 0.449\\ -0.060\\ 0.473\end{tabular} & \begin{tabular}[c]{@{}c@{}}-0.644\\ 0.288\\ 0.307\\ -0.602\\ 0.347\end{tabular} & \begin{tabular}[c]{@{}c@{}}-0.150\\ 0.494\\ 0.464\\ -0.179\\ 0.491\end{tabular}
\end{tabular}
\caption{Best silhouette scores for each combination of vectors. Numbers in bold indicate the best results for each column.\label{tab1}} 
\end{table*}
\label{table:res}
\endgroup
\begin{figure*}[]
\centering
\includegraphics[width=0.7\textwidth]{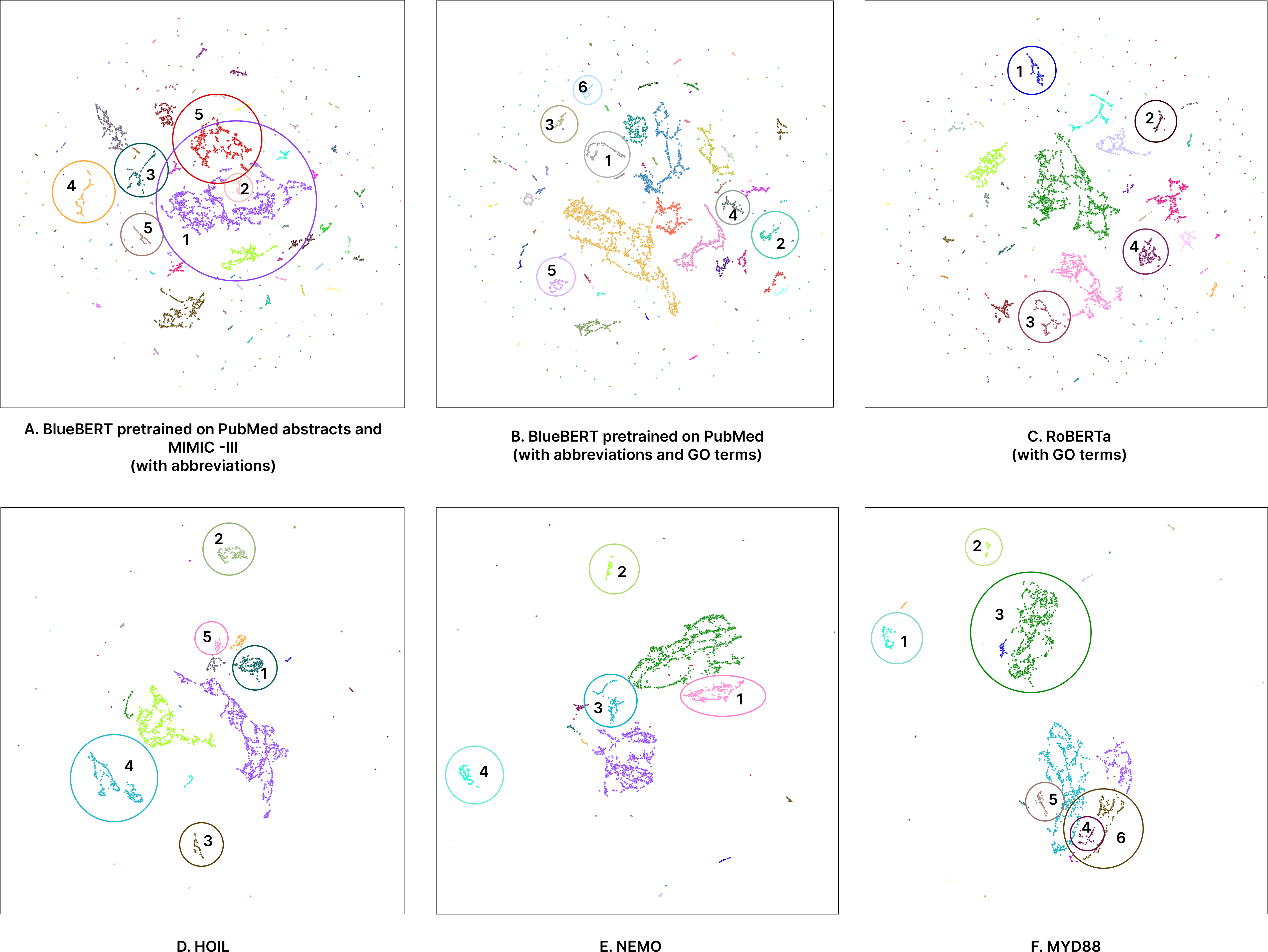}
\caption{The results created by the pipeline variants are shown in A-C. The plots D-F present clusterisation created by the variant with BlueBERT pretrained on PubMed with acronyms and GO-terms for three autoinflammatory diseases.}
\label{fig}
\end{figure*}
% that use the following three models: (A) BlueBERT pretrained on PubMed abstracts and MIMIC-II with abbreviations, (B) BlueBERT pretrained on PubMed with abbreviations and GO terms, (C) RoBERTa with GO terms

\section{Results}

\subsection{Quantitative estimation}

The Table ~\ref{table:res} contains the results of our computational experiments estimated by the silhouette score. Using this information, we have come to the following conclusions: BlueBERT pretrained on PubMed abstracts and MIMIC-III in combination with acronyms has shown the best score of 0.546 with UMAP as dimensional reduction method, and furthermore, it’s got 0.377 with T-SNE which also is the best score among all the combinations that use it. However, it has got the silhouette score notably worse than most of the other models in combination with GO-terms or GO-terms with acronyms. BlueBERT pretrained on PubMed abstracts, in contrast, has shown better results in these combinations, getting the highest score among the models. 

According to the obtained results, the three variant of the pipeline that require more attention are: BlueBERT pretrained on PubMed and MIMIC-III in combination with acronyms (\textbf{A}), BlueBERT pretrained on PubMed with acronyms and GO terms (\textbf{B}), and RoBERTa with GO terms (\textbf{C}). The expert has analyzed clusters created by each variant, and chose the ones that have an obvious biological interpretation.

\subsection{Expert analysis of custerization}

The three pipelines (\textbf{A},\textbf{B},\textbf{C}) proposed similar clusters of genes: transcriptional factors (see Fig.\ref{fig}A, 4-5; Fig.\ref{fig}B clusters 2-3; Fig.\ref{fig}C, 1), genes involved in trafficking processes and related cell components (endoplasmic reticulum, microtubules) (see Fig.\ref{fig}A, 6; Fig.\ref{fig}B, 1; Fig.\ref{fig}C, 3). Both pipelines that use BlueBERT models showed clusters of genes involved in the ubiquitination process (see Fig.\ref{fig}A, 3; Fig.\ref{fig}B, 6). Finally, pipelines \textbf{B} and \textbf{C} distinguished genes, that are encoding membrane proteins, or play a part in the maintenance of cell membrane (see Fig.\ref{fig}B, 1; Fig.\ref{fig}C, 3). The rest of clusters differed from one model to another.

The results of using the \textbf{A} pipeline, showed more easy-explainable clusters. For instance, a group of genes with the description of chemical reactions was found (see Fig.\ref{fig}B, 4). Another small distinguished cluster located on the left side (see Fig.\ref{fig}B, 5) contains genes that belong to either Rho or Rab small GTPases subgroups, which are known to interact with each other. The \textbf{C} pipeline has shown only 3 clusters which could be easily described, that were mentioned earlier, and the other didn't have many common features.

Also, there is additional expert opinion about the clusterization that was provided by various pipelines. Language models that are pretrained on medical texts show better performance than the models trained on general texts, as they provide more informative clusters. They catch and represent the meaning of gene functions more precisely. Adding GO-terms and acronyms to the clustered vectors helps to improve performance, because obtained clusters show not only functional similarity, but also different types of connections, such as involvement in similar signaling pathways, belonging to the same cell components, or to different subgroups of one gene family.

\subsection{Case of usage}

Aiming to present possible cases of the usage for the proposed approach, we use the pipeline \textbf{B} to list of genes for each disease separately. Clusters with similar common functions were found in every disease: groups of transcription factors (see Fig.\ref{fig}D, 3; Fig.\ref{fig}E, 1; Fig.\ref{fig}F, 3), genes responding to DNA damage (see Fig.\ref{fig}D, 2; Fig.\ref{fig}E, 4; Fig.\ref{fig}F, 1), and genes involved in transporting: ion channels, and aquaporins (see Fig.\ref{fig}D, 4; Fig.\ref{fig}E, 2; Fig.\ref{fig}F, 6). Both HOIL and NEMO diseases plots showed clusters with genes playing a part in immune response (see Fig.\ref{fig}D, 1; Fig.\ref{fig}E, 3). 

We managed to define other interesting clusters for each disease. In HOIL1 dataset a group of genes connected to the endoplasmic reticulum was found (see Fig.\ref{fig}D, 5). Its representatives participate in vesicle transporting from the  endoplasmic reticulum to Golgi complex, or from the plasmatic membrane to the endoplasmic reticulum. Some presented genes have a role in ERAD pathway, intended to perform clearance of misfolded proteins in the endoplasmic reticulum.

The next NEMO disease was examined. The distinct cluster at the top (see Fig.\ref{fig}F, 2) unites transcriptional factor genes. The second examined cluster (see Fig.\ref{fig}F, 1) had different membrane elements, including a big number of membrane ion channels. Once again, genes responding to DNA damage were found (see Fig.\ref{fig}F, 4). In addition, a cluster (see Fig.\ref{fig}F, 3) contained genes involved in vesicular trafficking, or immune response. This is important as a number of works stated that vesicular trafficking is essential for the secretion of immune regulators, involved in inflammatory reactions. % \cite{Melo2016}. 

MYD88 dataset revealed groups of genes which have chemical reactions in their description (see Fig.\ref{fig}F, 3). Another group includes Nf-kb activators, regulators of MAP signaling pathway, GTPases (see Fig.\ref{fig}F, 3). Notable, GTPases are involved in the Nf-kb regulation, which responds to inflammatory cytokines, and MAP signaling pathway plays a vital part during inflammation by producing pro-inflammatory cytokines.  

\section{Discussion and Conclusion}

After performing manual analysis, the best performing version of the pipeline was obtained. The steps are the following: text descriptions are encoded using BlueBERT pretrained on PubMed, GO-terms and acronyms are encoded in binary vectors (the \textbf{B} pipeline). After that dimensional reduction is performed with UMAP algorithm, all vectors are concatenated together and proceeded to the next dimensional reduction using the same method. The results are separated into clusters using k-means clustering method.

In the end, the proposed approach to the gene set analysis was considered to be useful. It can handle large lists of genes, reducing analysis of several thousand genes to analysis of several clusters.

\bibliographystyle{IEEEtran}
\bibliography{references}

\begin{thebibliography}{10}
\providecommand{\url}[1]{#1}
\csname url@samestyle\endcsname
\providecommand{\newblock}{\relax}
\providecommand{\bibinfo}[2]{#2}
\providecommand{\BIBentrySTDinterwordspacing}{\spaceskip=0pt\relax}
\providecommand{\BIBentryALTinterwordstretchfactor}{4}
\providecommand{\BIBentryALTinterwordspacing}{\spaceskip=\fontdimen2\font plus
\BIBentryALTinterwordstretchfactor\fontdimen3\font minus
  \fontdimen4\font\relax}
\providecommand{\BIBforeignlanguage}[2]{{%
\expandafter\ifx\csname l@#1\endcsname\relax
\typeout{** WARNING: IEEEtran.bst: No hyphenation pattern has been}%
\typeout{** loaded for the language `#1'. Using the pattern for}%
\typeout{** the default language instead.}%
\else
\language=\csname l@#1\endcsname
\fi
#2}}
\providecommand{\BIBdecl}{\relax}
\BIBdecl

\bibitem{venn}
A.~Jia, L.~Xu, and Y.~Wang, ``Venn diagrams in bioinformatics,''
  \emph{Briefings in Bioinformatics}, vol.~22, 04 2021.

\bibitem{pathways}
M.~A. García-Campos, J.~Espinal-Enríquez, and E.~Hernández-Lemus, ``Pathway
  analysis: State of the art,'' \emph{Frontiers in Physiology}, vol.~6, 2015.

\bibitem{pathways2}
P.~Khatri, M.~Sirota, and A.~Butte, ``Ten years of pathway analysis: Current
  approaches and outstanding challenges,'' \emph{PLoS computational biology},
  vol.~8, p. e1002375, 02 2012.

\bibitem{ppi}
C.~Xie, J.~Gao, Y.~Yuan, and Y.~Yu, ``Protein-protein interactions and their
  network analysis in bioinformatics,'' vol.~29, pp. 465--469, 04 2009.

\bibitem{GEOdataset}
Z.~Xu, ``Transcriptional analysis of whole blood in patients with
  auto-inflammatory disorders,'' 2012.

\bibitem{Stafford2007}
P.~Stafford and M.~Brun, ``\BIBforeignlanguage{eng}{Three methods for
  optimization of cross-laboratory and cross-platform microarray expression
  data},'' \emph{\BIBforeignlanguage{eng}{Nucleic acids research}}, vol.~35,
  no.~10, pp. e72--e72, 2007, 17478523[pmid].

\bibitem{bert}
J.~Devlin, M.-W. Chang, K.~Lee, and K.~Toutanova, ``Bert: Pre-training of deep
  bidirectional transformers for language understanding,'' 2018.

\bibitem{roberta}
Y.~Liu, M.~Ott, N.~Goyal, J.~Du, M.~Joshi, D.~Chen, O.~Levy, M.~Lewis,
  L.~Zettlemoyer, and V.~Stoyanov, ``Roberta: A robustly optimized bert
  pretraining approach,'' 2019.

\bibitem{clinicalBERT}
E.~Alsentzer, J.~Murphy, W.~Boag, W.-H. Weng, D.~Jindi, T.~Naumann, and
  M.~McDermott, ``Publicly available clinical {BERT} embeddings,'' in
  \emph{Proceedings of the 2nd Clinical Natural Language Processing
  Workshop}.\hskip 1em plus 0.5em minus 0.4em\relax Minneapolis, Minnesota,
  USA: Association for Computational Linguistics, Jun. 2019, pp. 72--78.

\bibitem{bioBERT}
J.~Lee, W.~Yoon, S.~Kim, D.~Kim, S.~Kim, C.~So, and J.~Kang, ``Biobert: a
  pre-trained biomedical language representation model for biomedical text
  mining,'' \emph{Bioinformatics (Oxford, England)}, vol.~36, 09 2019.

\bibitem{blueBERT}
Y.~Peng, S.~Yan, and Z.~lu, ``Transfer learning in biomedical natural language
  processing: An evaluation of bert and elmo on ten benchmarking datasets,'' 06
  2019.

\bibitem{umap}
L.~McInnes, J.~Healy, and J.~Melville, ``Umap: Uniform manifold approximation
  and projection for dimension reduction,'' 2018.

\bibitem{clustering}
J.~MacQueen \emph{et~al.}, ``Some methods for classification and analysis of
  multivariate observations,'' in \emph{Proceedings of the fifth Berkeley
  symposium on mathematical statistics and probability}, vol.~1, no.~14.\hskip
  1em plus 0.5em minus 0.4em\relax Oakland, CA, USA, 1967, pp. 281--297.

\bibitem{sil_score}
P.~Rousseeuw, ``Rousseeuw, p.j.: Silhouettes: A graphical aid to the
  interpretation and validation of cluster analysis. comput. appl. math. 20,
  53-65,'' \emph{Journal of Computational and Applied Mathematics}, vol.~20,
  pp. 53--65, 11 1987.

\end{thebibliography}

\vspace{12pt}
\end{document}